# Time series Forecasting to detect anomalous behaviours in Multiphase Flow Meters

T. Barbariol, Università degli Studi di Padova
D. Masiero, Università degli Studi di Padova
E. Feltresi, Pietro Fiorentini
G.A. Susto, Università degli Studi di Padova

**INTRODUCTION**

Multiphase flow meters (MPFM) are inline metering tools, able to measure the individual flow rates of a multiphase flow. These instruments measure multiple properties of the fluid and then compute the water, oil and gas flow rates. These derived quantities are then fed into reservoir models in order to make predictions on the future production, or to allocate revenues and taxes. The main fluid properties measured by the MPFM are the electrical impedance, the gamma-ray absorption, and the velocity of the flow. These are sampled by means of three modules: electrical probes, gamma-ray densitometer and Venturi tube.

The reliability of these modules is of primary importance: if one of them experiences a fault, the provided flow measurements will be incorrect. This might affect in cascade all the following steps, leading to wrong decisions about the reservoir [1]. Without proper supervision, one module might fail without anyone noticing it and without expert human intervention.

In order to prevent these situations, a system able to detect anomalous behaviours is needed: this system has to work autonomously, and to adapt to different working conditions without losing its efficacy. These intelligent modules go under the name of Anomaly Detection (AD), or, more in general, Fault Detection (FD) systems. They are applied to multiple objects that have strong reliability requirements, such as: smart grids [2], medical devices [3], autonomous vehicles [4]. Depending on the approach, AD/FD technologies can be divided into two categories: model-driven or data-driven approaches. Due to its high flexibility and effectiveness in this work we described the adoption of data-driven approaches in the field of MPFM.

This is one of the first works that addresses the fault detection problem in Multiphase Flow Meters, employing time-series forecasting tools. Moreover, this is one of the earliest applications of the Temporal Convolutional Network to time-series anomaly detection.

This paper is structured as follows: in Section 2 the MPFM fault detection problem addressed in this paper will be described in more detail, together with the algorithmic tools employed in this analysis. Then in Section 3 these tools will be applied to real faults. In the last Section the conclusions of this paper will be discussed.

## 2   MATERIALS AND METHODS: MPFM and ML

### 2.1   Multiphase flow meters

As already mentioned in the introduction, the MPFM is a complex metrology instrument, able to quantify the individual flow rates of a mixture. It is placed on the top of an oil well to measure the oil and gas production. Since the reservoir has an evolution in time, also the MPFM experiences a variation of its measures during the reservoir lifetime. It makes continuous measurements of the fluid properties, in particular electrical impedance (C) and gamma-ray absorption (G). Keep in mind that these two sensors might not be placed at the same point of the pipeline, and therefore their measurements might not be aligned in time. This apparent disadvantage can be seen as an advantage when one wants to predict the behaviour of the subsequent sensor (in this case *G*) given the precedent sensor (*C*). The method explained in the following paragraphs will be applied to these two sensor measurements since they exhibit a quite high correlation at a small time scale.

## 2.2   Fault Detection

The proposed approach is widely adopted in many industrial applications [5,6]. Since sensor measurements are continuous, the most natural way to tackle the fault detection problem is by using time series-based approaches.

The one used in this paper consists of three steps: (i) forecasting, (ii) residue analysis and (iii) fault alarm. The main idea behind it is:
-       forecasting: at each time step *t* the following value at time *t+k* is predicted based on values in the time interval *[t-n,t]*, where k is the number of steps ahead, and n is the size of the selected time interval.
-       residue analysis and fault alarm: at time *t+k*, if the difference between the predicted value and the actual value is too high, the actual value is considered anomalous.
In the following paragraphs, this simple idea will be better formalised and explained.

**Time Series Forecasting**

There exist two types of time-series forecasting: the short-term forecasting when the model tries to predict only one step ahead (*k=1*), and the long-term forecasting when the model forecasts multiple steps ahead (*k>1*). This last approach might be less computationally intensive, but it is for sure less accurate than the short-term forecasting. For this reason in this paper the choice was to adopt a one one-step ahead forecasting.

In recent years many models were developed for this kind of task. A particular branch of Machine Learning, named Deep Learning, is proving its superiority in many fields like computer vision and natural language processing. The most promising model, developed for time series forecasting, is the Temporal Convolutional Network TCN [7,8]. In the present work, the authors decided to test this tool with two configurations, the *endogenous* and *exogenous* one. The first has the same input and output signal (*G*) while the second has different signals (in the following example, *C* as input and *G* as output).

Before going any further, it is better to fix the notation: *x(t)* represents the input signal, while *y(t)* the output one. A generic model takes a batch of input values *[x(t-n),x(t-n-1),...,x(t-1),x(t)]* and gives back the short-term forecast *y(t+1)*.

Two additional models were tested to assess the goodness of the TCN performances.
The first, called Naive forecasting, is a common baseline: it predicts the value at *t+1* as the value collected at time *t*. The second, named Hard Subtraction, follows the approach originally developed in [9]: one signal is approximated by the *time-aligned* version of another, correlated signal. If *y* and *x* are misaligned by *m* time steps, then the forecast will be *y(t+1)=x(t+1-m)*.

**Residue analysis and fault alarm**

Once the forecast has been obtained, the next step is the analysis of the residue between the forecast and the actual value. Obviously, the higher this difference is, the higher will be the probability that the actual value is indeed anomalous. A threshold on the residue is needed in order to discriminate if one value has to be flagged as anomalous or not. This threshold can be learned in advance, by looking at some statistical properties of the past residues.

More specifically, the procedure followed in this paper is the following. Since the residue has a quite strong fluctuation, a time-window has been defined and the instantaneous residue has been substituted by two derived quantities: the absolute rolling mean (*mean*) and the absolute rolling standard deviation (*std*). The thresholds for these quantities have been learnt in advance, when the MPFM was working in controlled conditions. Each threshold corresponds to the maximum value assumed by its corresponding quantity in this specific time interval.

When the *mean* or the *std* exceeds its threshold, the fault is detected and an alarm is raised.

## 3    RESULTS

In this Section, the results of the simulations will be described, starting from the forecast, passing through the residue analysis, and ending with real and synthetic fault tests that will prove the effectiveness of the proposed method.

Two signals have been collected: the electric impedance of the flow *C*, and the gamma-ray attenuation *G*. For simplicity the target signal will be G, but for construction an anomaly on G will trigger an anomaly C too and vice versa.

### 3.1    Forecast

The metric that shows the general goodness of the forecast is depicted in Figure 1. It is the mean squared error (*mse*), a metric commonly used in ML to assess the performance of a model. Lower it is, the better the model is performing. What is important is the *mse* value of forecasting models, compared to the Naive forecasting (the baseline).

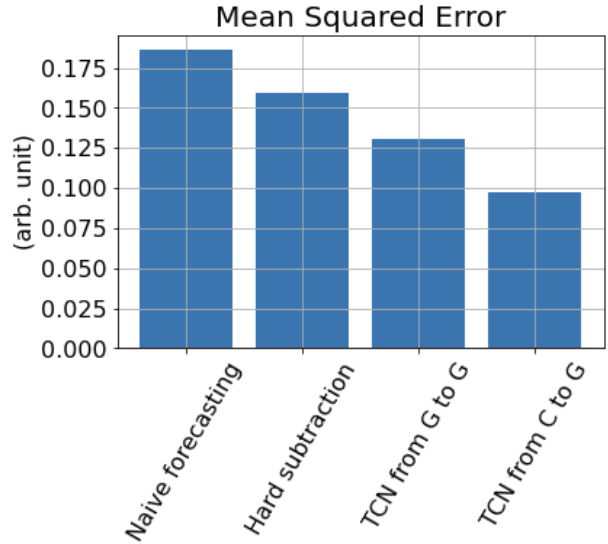

Figure 1: Comparison of the forecasting performances.

All the methods perform better than the baseline, indicating that the training procedure was likely appropriate. As expected, the TCN that maps *C* to *G* outperforms the other approaches. Despite Hard Subtraction performs slightly worse than TCN based on the signal *G*, in the next paragraph a more solid perspective will be embraced.

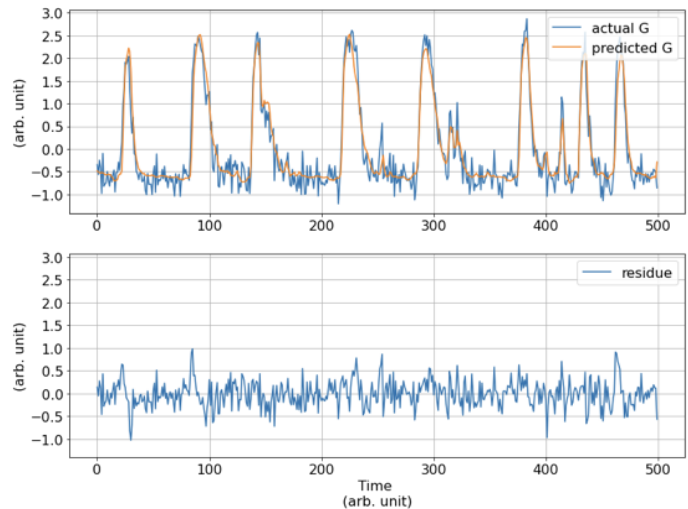

Figure 2: Forecasting example.

In Figure 2, an example of the TCN performance is shown. It can be seen the great ability of the model to accurately predict the high values of the signals. Below the forecast, the approximately gaussian residue is depicted: smoothing the fluctuations with a moving average would lead to a null residue.

### 3.2 Robustness of the forecast

As previously anticipated, not only the forecast precision is important in this context. In Figure 3 two TCNs are trained to forecast the value of signal *G* exposed to a fault around the time step 100. In the first Figure the TCN was trained over the same signal *G* while in the second Figure the TCN was trained using *C* as input.

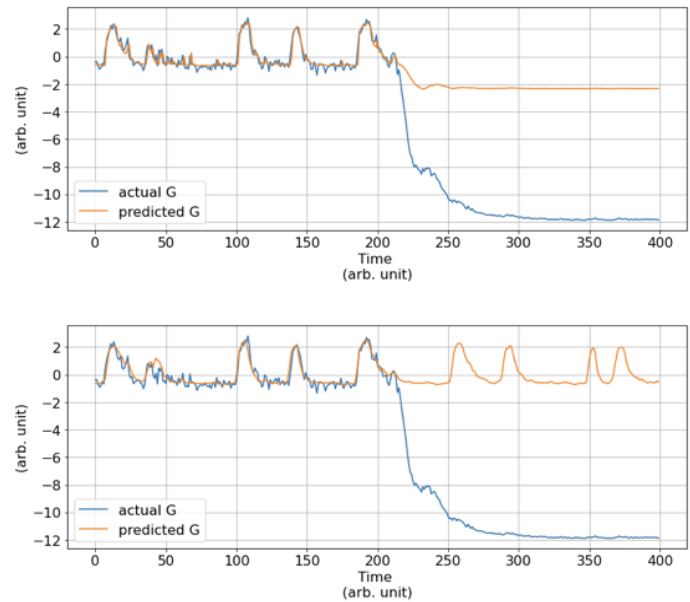

Figure 3: Comparison between the forecast of the TCN trained on endogenous and exogenous variables.

In the first case, when the model is trained on the endogen variable *G*, the forecast adapts to the fault and it is not able to highlight it. On the contrary, when the model is trained on the exogen variable *C*, the forecast is not affected by the fault on G. For this reason, hard subtraction might be preferable in contrast to the TCN trained on endogenous variables.

### 3.4  Faults

In the following paragraphs, the proposed approach for MPFM fault detection will be tested over synthetic and real faults. The model used in the forecast is the TCN with *C* in input and *G* as output.

**Synthetic Faults**
Generating synthetic faults allows simulating the faults too difficult or expensive to create in a laboratory. Four kinds of them have been simulated: a complete failure of the sensor, a degradation of the sensor accuracy, a drift and a sudden bias in the sensor measurement.
The alarm thresholds have been learnt in the very first time intervals.

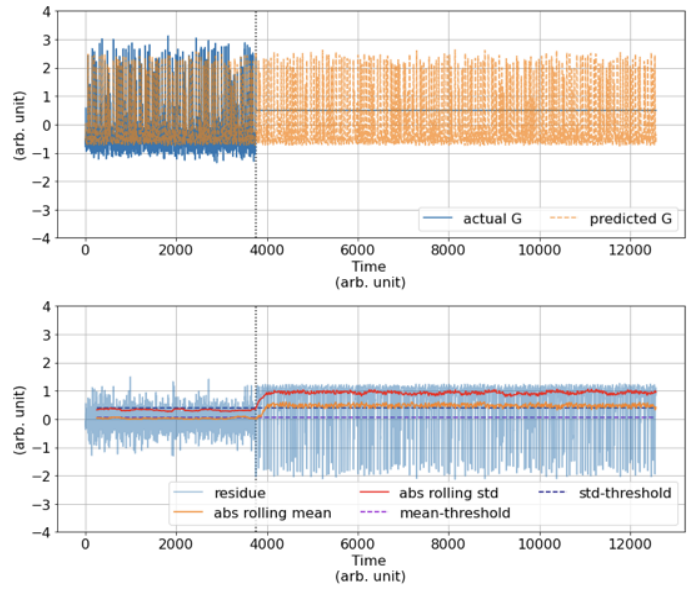

Figure 4: Complete failure fault.

In Figure 4 the complete failure is plotted along with its residue. As already shown, the forecast is not affected by the fault. In this case both the *mean* and *std* exceed the learnt thresholds causing a fault alarm triggering.

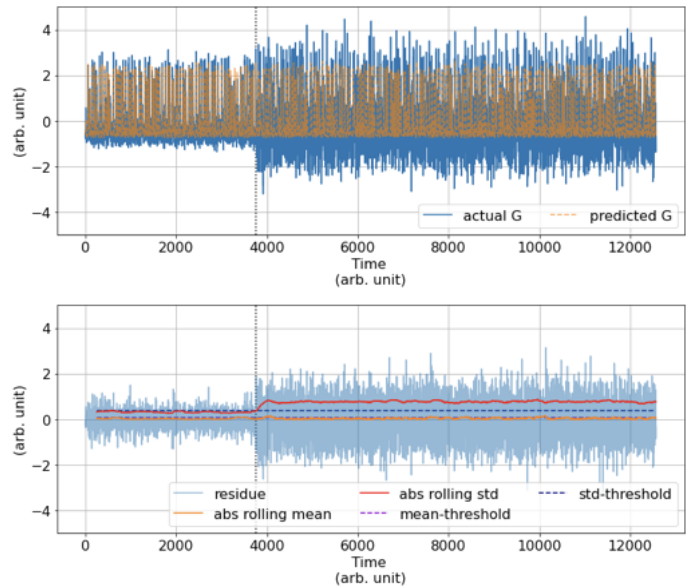

Figure 5: Precision degradation fault.

Figure 5 shows a different kind of fault: the precision degradation. In this case the mean residue remains zero, but the std increases and triggers the detection of the fault.

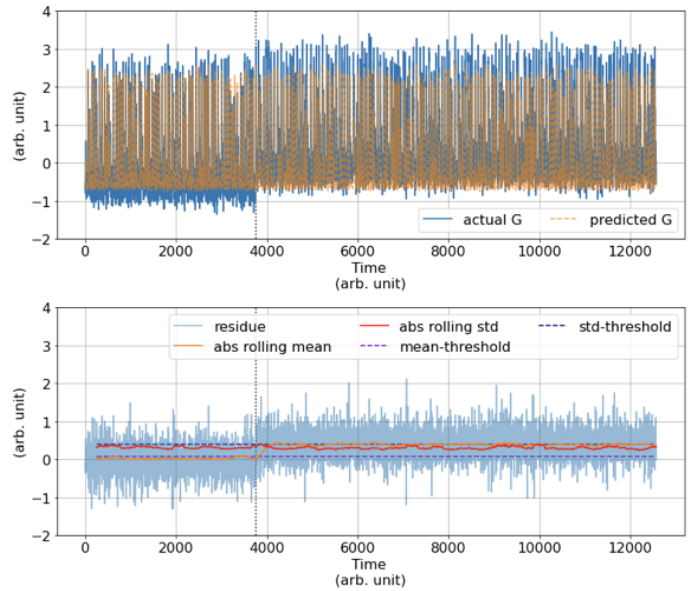

Figure 6: Bias fault.

On the contrary, the bias fault visible in Figure 6 is detected by the sudden change of mean.

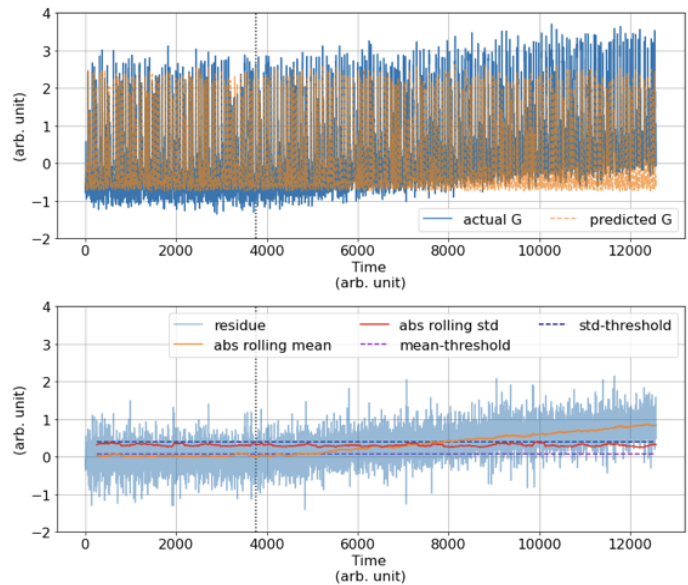

Figure 7: Linear drift fault.

The drift fault is more subtle and difficult to detect (Figure 7). The *std* remains constant and the mean increases very slowly therefore this fault type can be confused with a natural change of MPFM working condition.

### 3.4 Real Faults

In this section the application of the proposed approach to real faults is shown. Real faults were generated by a human operator, disconnecting cables and closing valves of the MPFM, simulating what it can experience in the most common real operating conditions.

**Closed Gamma-ray shutter**

In this case, the shutter of the gamma-ray densitometer was closed. This causes a sudden drop in the measured density that leads to a wrong estimation of the mass flowing in the pipe (Figure 8). The residue raises very quickly and the absolute rolling mean exceeds its threshold, triggering the alarm. Here, the absolute rolling standard deviation has a little contribution in the fault detection.

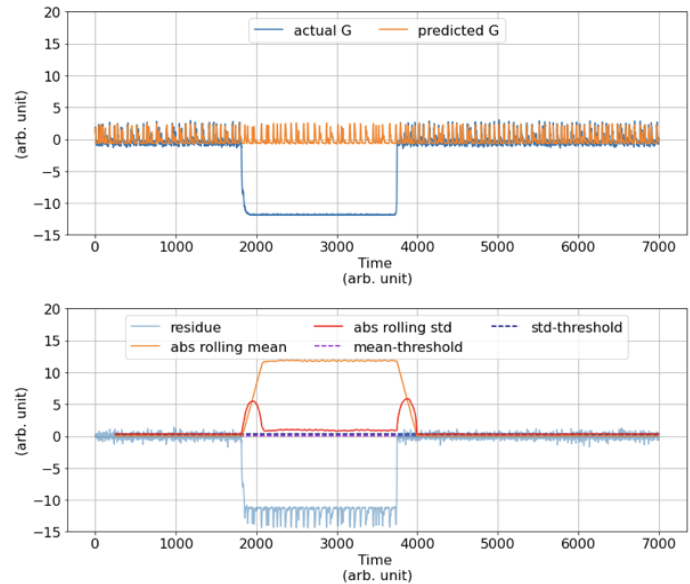

Figure 8: Closed Gamma-ray shutter.

**Unplugged communication cable**

Although the closed shutter could have been detected by a much simpler method, the disconnected communication cable is much more complex. In this case the electronics continue to receive the last recorded sequence over and over again. Note that at the time when this sequence was firstly recorded, it was normal, so a univariate control method can hardly detect this type of anomaly.

Looking at (Figure 9) it is quite difficult to see the generated fault, but fortunately the proposed method promptly detects it. The exploding residue is quite clear and the rolling std raises quickly.

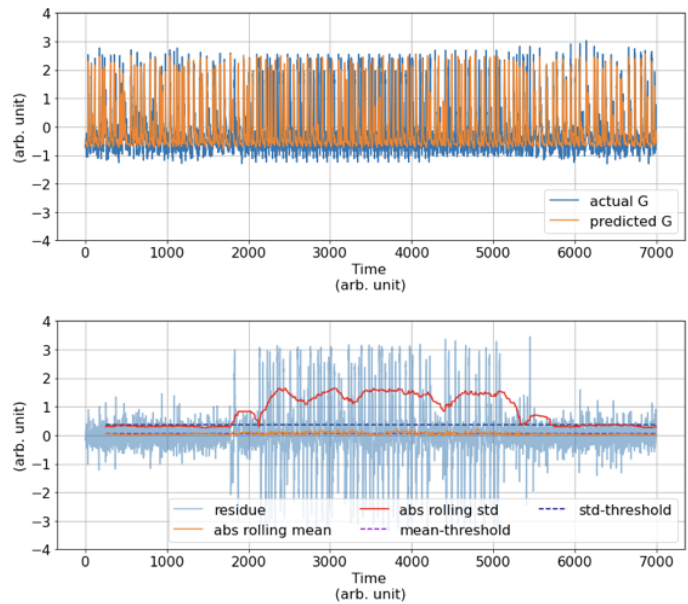

Figure 9: Unplugged communication cable.

## 4 CONCLUSIONS

In this paper an approach to the fault detection of Multiphase Flow Meters has been proposed. It applies to its sensors and specifically to its time-series measures. It consists of three steps: forecast, residual analysis and fault alarm.
In the forecasting step, care has been taken not only to the prediction accuracy, but also to the robustness of the method. After the development, it was applied to real faults and it proved its ability to easily detect the generated faults.

The authors see two natural research directions for future works: to test new prediction models trying to improve the predictive performances, and to make long-term forecasts in order to lessen the computational burden of the method.

## 5 REFERENCES

[1]     L.S HANSEN, S. PEDERSEN, P. DURDEVIC. "Multi-phase flow metering in offshore oil and gas transportation pipelines: Trends and perspectives." *Sensors* 19.9 (2019): 2184.

[2]     C.A. ANDRESEN, B.N. TORSAETER, H. HAUGDAL, K. UHLEN. "Fault detection and prediction in smart grids." *2018 IEEE 9th International Workshop on Applied Measurements for Power Systems (AMPS)*. IEEE, 2018.

[3]     L. MENEGHETTI, M. TERZI, S. DEL FAVERO, G.A. SUSTO, C. COBELLI " Data-driven anomaly recognition for unsupervised model-free fault detection in artificial pancreas." *IEEE Transactions on Control Systems Technology* 28.2, pag. 33-47 (2020).

[4]     J. REN, R. REN, M. GREEN, X. HUANG. "A deep learning method for fault detection of autonomous vehicles." *2019 14th International Conference on Computer Science & Education (ICCSE)*. IEEE, 2019.

[5]     S. AHMAD, S. PURDY. "Real-time anomaly detection for streaming analytics." *arXiv preprint arXiv:1607.02480* (2016).

[6]     K. HUNDMAN, V. CONSTANTINOU, C. LAPORTE, I COLWELL, T. SODERSTROM. "Detecting spacecraft anomalies using lstms and nonparametric dynamic thresholding." *Proceedings of the 24th ACM SIGKDD international conference on knowledge discovery & data mining*. 2018.

[7]     S. BAI, J.Z. KOLTER, V. KOLTUN. "An empirical evaluation of generic convolutional and recurrent networks for sequence modeling." *arXiv preprint arXiv:1803.01271* (2018).

[8]     Y. JINING, M. LIN, L. WANG, R. RAJIV, A.Y. ZOMAYA. "Temporal Convolutional Networks for the Advance Prediction of ENSO." *Scientific Reports (Nature Publisher Group)* 10.1 (2020).

[9]     T. BARBARIOL, E. FELTRESI, G.A. SUSTO. "Self-Diagnosis of Multiphase Flow Meters through Machine Learning-Based Anomaly Detection." *Energies* 13.12 (2020): 3136.